\documentclass[sigconf]{acmart}

\usepackage{subfigure}
\usepackage[ruled,vlined]{algorithm2e}
\usepackage{array, makecell} 

\usepackage{acmart-taps}

\usepackage{amsmath}

\AtBeginDocument{%
  \providecommand\BibTeX{{%
    \normalfont B\kern-0.5em{\scshape i\kern-0.25em b}\kern-0.8em\TeX}}}
\copyrightyear{2025}
\acmYear{2025}
\setcopyright{cc}
\setcctype{by}
\acmConference[ICAAI 2025]{2025 9th International Conference on Advances in Artificial Intelligence}{November 14--16, 2025}{Manchester, United Kingdom}
\acmBooktitle{2025 9th International Conference on Advances in Artificial Intelligence (ICAAI 2025), November 14--16, 2025, Manchester, United Kingdom}
\acmDOI{10.1145/3787279.3787282}
\acmISBN{979-8-4007-2104-5/2025/11}

\begin{document}

\title{Evaluation of Bagging Predictors with Kernel Density Estimation and Bagging Score}

\author{Philipp Seitz}
\authornote{Corresponding Author}
\email{philipp.seitz@thws.de}
\orcid{0009-0002-1096-1721}
\affiliation{%
  \institution{Institute of Digital Engineering, Technical University of Applied Sciences W\"{u}rzburg-Schweinfurt}
  \streetaddress{Ignaz-Schön-Straße 11}
  \city{Schweinfurt, Bavaria}
  \country{Germany}
  \postcode{97421}
}

\author{Jan Schmitt}
\email{jan.schmitt@thws.de}
\orcid{0000-0003-4537-7680}
\affiliation{%
  \institution{Institute of Digital Engineering, Technical University of Applied Sciences W\"{u}rzburg-Schweinfurt}
  \streetaddress{Ignaz-Schön-Straße 11}
  \city{Schweinfurt, Bavaria}
  \country{Germany}
  \postcode{97421}
}

\author{Andreas Schiffler}
\email{andreas.schiffler@thws.de}
\orcid{0000-0002-1447-7331}
\affiliation{%
  \institution{Institute of Digital Engineering, Technical University of Applied Sciences W\"{u}rzburg-Schweinfurt}
  \streetaddress{Ignaz-Schön-Straße 11}
  \city{Schweinfurt, Bavaria}
  \country{Germany}
  \postcode{97421}
}

\begin{abstract}
For a larger set of predictions of several differently trained machine learning models, known as bagging predictors, the mean of all predictions is taken by default. Nevertheless, this proceeding can deviate from the actual ground truth in certain parameter regions. An approach is presented to determine a representative value $\Tilde{y}_{BS}$ from such a set of predictions using Kernel Density Estimation (KDE) in nonlinear regression with Neural Networks (NN) which simultaneously provides an associated quality criterion $\beta_{BS}$, called Bagging Score (BS), that reflects the confidence of the obtained ensemble prediction. It is shown that working with the new approach better predictions can be made than working with the common use of mean or median. In addition to this, the used method is contrasted to several approaches of nonlinear regression from the literatur, resulting in a top ranking in each of the calculated error values without using any optimization or feature selection technique.
\end{abstract}




\maketitle

\section{Introduction}
\label{section_introduction}

Machine Learning (ML) is generally used, for example, to recognize objects in images and assign them to predefined groups, called classification~\cite{hunter2012selection}, or to best approximate an unknown function using given data points, called Regression~\cite{sykes1993introduction}. 
Nevertheless, for high prediction accuracy, large data sets are basically needed that can be used to train ML models. The more extensive the data, the better the predictions. It especially becomes problematic when small data sets are available. In this case, so-called bagging (\textbf{b}ootstrap \textbf{agg}regat\textbf{ing}, introduced by~\cite{breiman1996bagging}) brings a possible solution. By splitting the available data into different training and validation sets, several models can be trained differently with the same dataset, so called boosting~\cite{schapire2003boosting}~\cite{drucker1994boosting}, but do not make unanimous conclusions in subsequent predictions. For decision making in classification tasks, for example, the majority formation of the ensemble is then used, which can also still be used for integer outputs in the regression~\cite{wang2010high}~\cite{bauer1999empirical}.

Though finding a majority proves difficult when outputs of the networks are generated as decimal numbers. In this case, the literature considers the calculation of the mean or median~\cite{friedman2007bagging}~\cite{chen2009bagging}~\cite{wang2010high}~\cite{bauer1999empirical}~\cite{grandvalet2004bagging}. 

The concept of bagging is widely used in the literature and achieves consistently better results.~\cite{guo2022predicting} analysed the predictability of soil liquefaction potential between several machine learning classification models where bagging achieved the best results.~\cite{lin2024comprehensive} investigated the performances between bagging and classical single predictive models which concluded that bagging outperforms them all.

\begin{figure}[htbp]
\centering
\subfigure[Expected Distribution]{%
\resizebox*{6cm}{!}{\includegraphics{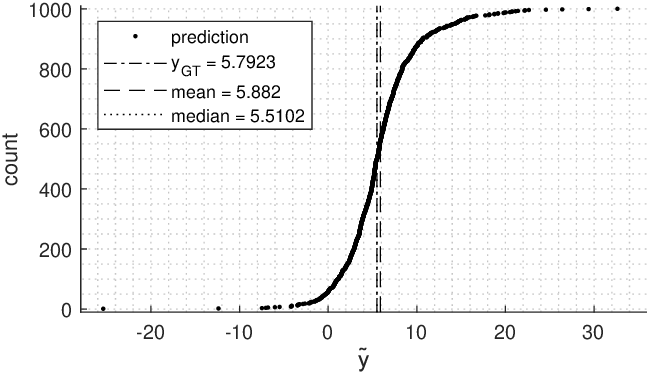}}
        \label{subfig_expectedDistribution}
        }\hspace{5pt}
\subfigure[Actual Common Distribution]{%
\resizebox*{6cm}{!}{\includegraphics{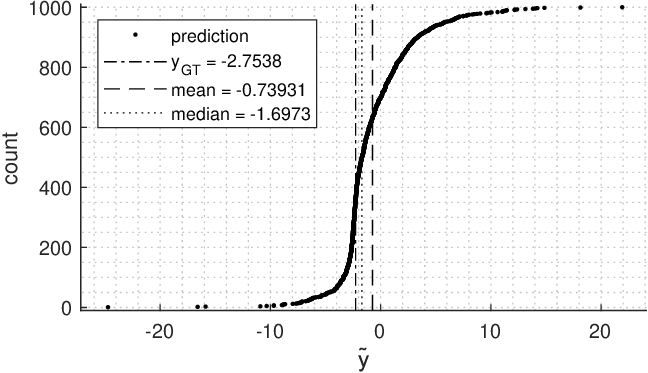}}
        \label{subfig_unexpectedDistribution}
        }\hspace{5pt}
\caption{Example for expected and unexpected usual asymmetric cumulative distribution when evaluating bagging predictors where mean and median are likely to deviate from the actual ground truth.}
\label{fig_goodbadDistribution}
\end{figure}

For the evaluation with mean or median, though, there is the problem that these procedures are susceptible to asymmetries in the distribution pattern of the statements (see Figure~\ref{fig_goodbadDistribution}).

One way to evaluate a large set of predictions more precisely is to use Kernel Density Estimation~\cite{parzen1962estimation}. It can be used to determine the density function of a set of points, which in the case of normally distributed random variables forms the Gaussian-curve. Depending on the shape and properties of the set of points, a deformed curve is obtained, which provides information about their asymmetric density distribution \cite{wkeglarczyk2018kernel}.


However, there is no known approach that focuses in particular on the evaluation of bagging predictors by using KDE to tackle the susceptance to asymmetrics in the distribution pattern using mean or median. 

In this paper, nonlinear regression is used to introduce a method for determining a representative majority score from a larger number of predictions and the associated so-called Bagging Score. It indicates how trustworthy the representative majority value is in a range between zero and one.

The application of the Bagging Score provides higher prediction accuracies from predictions of ML model ensembles compared to the use of standard statistical methods, which are contrasted in the course of this work.

Section~\ref{section_datasets} here describes the datasets used and Section~\ref{section_bagging_score} presents the method by which the representative statement of a set of NN predictions and the associated BS are obtained. In Section~\ref{section_comparative_analysis} the presented method is compared to the previously known approaches from the literature and to other methods from the literature with a real life dataset. Section~\ref{section_conclusion_and_further_research} closes the study with a conclusion and further research.

\section{Bagging Score}
\label{section_bagging_score}

Section~\ref{section_bagging_score} deals with the presentation of the procedure for determining the representative statement of the ensemble and their associated Bagging Scores for estimating the associated significance.

\subsection{Design of the Bagging Predictors}
\label{subsection_bagging}

The dataset is used for the training of $n_c=1000$ NNs (ensemble) consisting of $3$ hidden layers with $20$ neurons each, which later form a set of predictions $\Tilde{\mathcal{Y}_x}\in \mathbb{R}^{n_c\times 1}$ for a given point $x$ as the input of the NNs. Thereby, a random seed $R=i$ is set to initialize the network parameters and generate the training and validation set of the $i$-th network. The second hidden layer of the respective networks is equipped with the nonlinear transfer function $tanh(x)$. 

\begin{equation}
    tanh(x) = \frac{2}{1+e^{-2x}}-1
\end{equation}

The remaining hidden layers stay linear $\sigma_{nl}(x) = x$. Additional information on the functionality and choice of this constellation of transfer functions can be found in~\cite{seitz2023alternating}. The split into training and validation set happens randomly per network training in a ratio of $70\%$ to $30\%$.
For each set of predictions $\Tilde{\mathcal{Y}_x}$, a representative value $\Tilde{y_\gamma}(\Tilde{\mathcal{Y}_x})$ is formed using the respective statistical evaluation method $\gamma$ (MEAN and MEDIAN) and the presented method of this section, which together reflect the prediction of the ensemble.\\

\subsection{Methodology}
\label{subsection_methodology}

In general, the distribution of predictions should follow a normally distributed pattern, since most networks are statistically close to one specific value in their predictions and few over- and underestimate the actual value in an equally distributed manner. The mean or median could thus be used to give a good, representative prediction value. 
However, mean or median values do no longer necessarily match if predictions are especially made in unknown parameter areas, which where not in the training set. On the one hand, strong deviations can occur, if the distribution becomes more comparable with a logarithmic normal distribution. On the other hand, several different majority clusters can form, which in turn can have a significant influence on the mean and median formation (see Figure~\ref{fig_goodbadDistribution}).

An adapted method based on the KDE for normally distributed random variables should remedy this situation~\cite{chen2017tutorial}~\cite{sheather2004density} and gives a detailed insight into the distribution of the predictions. 

Using a kernel function $K(x,x_{\mu})$ (Eq.~\ref{eq_kdeFunction}) in the form of a Gaussian curve, the data points of the predictions $\Tilde{\mathcal{Y}}$ are summed weighted successively in a local interval of size $\sigma_{\Tilde{\mathcal{Y}}} =std(\Tilde{\mathcal{Y}})$ with a step size $\Delta \Tilde{y}$ from $\Tilde{y}_{start}=\min( \Tilde{\mathcal{Y}}) -\sigma_{\Tilde{\mathcal{Y}}}$ to $\Tilde{y}_{end}=max( \Tilde{\mathcal{Y}}) +\sigma_{\Tilde{\mathcal{Y}}}$ over the entire set of predictions. The kernel function is designed in such way that the maximum of the resulting density function (Bagging Score $\beta_{BS}$) for completely identical predictions is $\beta_{BS}=1$. (e.i. normalization factor  $\frac{1}{\sigma \sqrt{2\pi }}\hat{=}1$). The width $h_K$ of the kernel function is set to have its $99.7\%$ confidence interval at the current position $x_{\mu}$ at $[ x_{\mu}-\sigma_{\Tilde{\mathcal{Y}}}/2;x_{\mu}+\sigma_{\Tilde{\mathcal{Y}}}/2]$.

\begin{equation}
    K(x,x_{\mu})=\exp(-\frac{(x-x_{\mu})^2}{2\cdot h_k^2})
    \label{eq_kdeFunction}
\end{equation}
with\\
$h_k=\sigma_{\Tilde{\mathcal{Y}}} /6$\\

The overall procedure for estimating the density function is as follows:\\

\begin{algorithm}
 \KwData{Ensemble of predictions $\Tilde{\mathcal{Y}}$ from inputparameter $\Vec{x}$}
 \KwResult{Estimated Density Function $\mathcal{F}_{\Tilde{\mathcal{Y}}}$}
 $\Tilde{y}_{\mu}=\Tilde{y}_{start}$\;
 $\Delta \Tilde{y}=\frac{max(\Tilde{\mathcal{Y}})-min(\Tilde{\mathcal{Y}})}{1000}$\;
 \While{$\Tilde{y}_{\mu}\leq \Tilde{y}_{end}$}{
 $Y = \{ y | y \geq \Tilde{y}_{\mu}-\sigma_{\Tilde{\mathcal{Y}}}/2 \And y \leq \Tilde{y}_{\mu}+\sigma_{\Tilde{\mathcal{Y}}}/2\}\subseteq \Tilde{\mathcal{Y}}$\;
 $\mathcal{F}_{\Tilde{\mathcal{Y}}}(\Tilde{y}_{\mu})=\frac{1}{|\Tilde{\mathcal{Y}}|}\cdot \sum_{y\in Y}{K(y,\Tilde{y}_{\mu})}$\;
 $\Tilde{y}_{\mu}=\Tilde{y}_{\mu}+\Delta \Tilde{y}$;
 }
 \caption{Determining the Estimated Density Function $\mathcal{F}_{\Tilde{\mathcal{Y}}}$}
\end{algorithm}

\vspace{8mm}

\begin{figure*}[htbp]
\centering
\subfigure[Ideal: Evaluation of the Bagging Score at $f_{GT}(-10.73)$]{%
\includegraphics[width=1.0\textwidth]{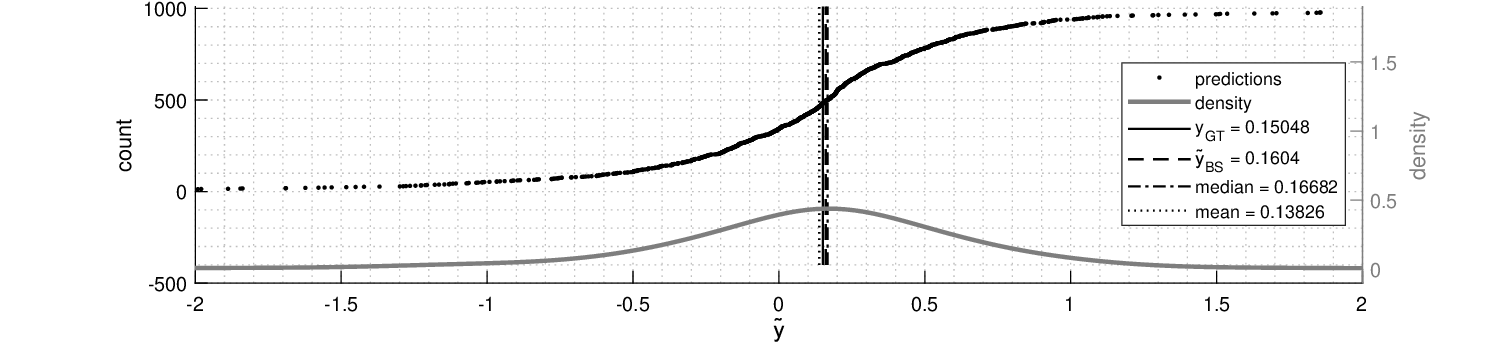}}
        \label{subfig_useCaseBS_ideal}
        \vspace{5pt}
\subfigure[Shift: Evaluation of the Bagging Score at $f_{GT}(-5.91)$]{%
\includegraphics[width=1.0\textwidth]{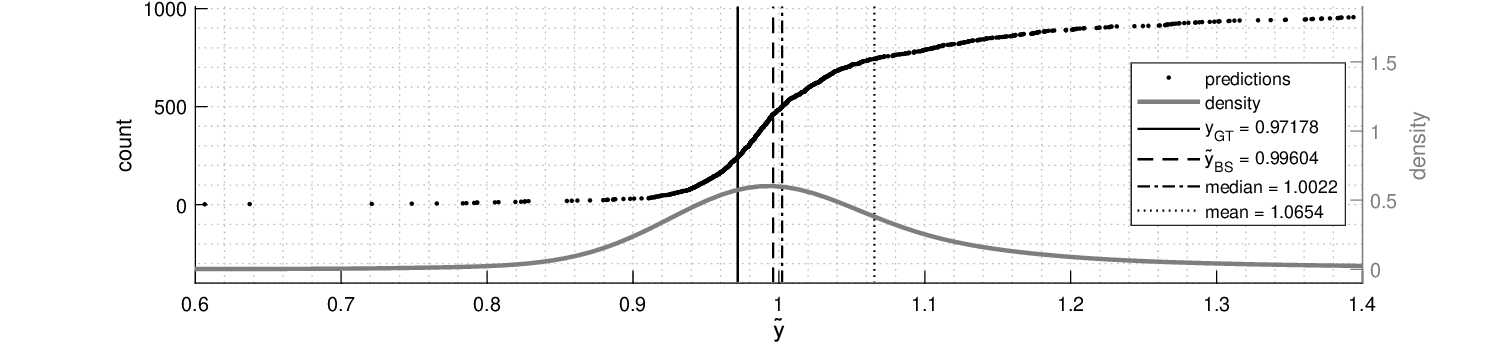}}
        \label{subfig_useCaseBS_nice}
        \vspace{5pt}
\caption{Use case of determining the ensemble prediction and its Bagging Score by two examples of a synthetic function $f_{GT}(x)$. Top: Example for good normal distribution in prediction set. Bottom: Example for shifted distribution.}
\label{fig_usecaseBaggingScore}
\end{figure*}

The y-coordinate of the maximum of the resulting density function $\mathcal{F}_{\Tilde{\mathcal{Y}}}(x)$ represents the Bagging Score $\beta_{BS}(\Tilde{\mathcal{Y}})$ whose x-coordinate denotes the representative prediction $\Tilde{y}_{BS}(\Tilde{\mathcal{Y}})$ of the ensemble.\\ 

\begin{equation*}
    \beta_{BS}(\Tilde{\mathcal{Y}}) = \max\{\mathcal{F}_{\Tilde{\mathcal{Y}}}(x)|\Tilde{y}_{start}\leq x \leq \Tilde{y}_{end}\}
\end{equation*}

\begin{equation*}
    \Tilde{y}_{BS}(\Tilde{\mathcal{Y}}) = \mathcal{F}^{-1}_{\Tilde{\mathcal{Y}}}(\beta_{BS}(\Tilde{\mathcal{Y}}))
\end{equation*}

Decreasing the step size $\Delta \Tilde{y}$ increases the resolution of the resulting density function. It is heuristically set to $\frac{1}{1000}th$ of the total interval of all predictions.\\

Figure~\ref{fig_usecaseBaggingScore} shows the outcome of an application to determine $\beta_{BS}$ and the corresponding $\Tilde{y}_{BS}$ of the predictions $\Tilde{\mathcal{Y}}$ from a trained ensamble using synthetic data. It is well seen that the mean can have a good alignment to the median and $\Tilde{y}_{BS}$ (top), but can also deviate strongly from the actual ground truth value, whereas the deviation by median and $\Tilde{y}_{BS}$ is much smaller (bottom).

It should be noted that a sufficient size of the ensemble is chosen, otherwise statistical inaccuracies will occur.

\section{Comparative Analysis}
\label{section_comparative_analysis}

In the following, the presented method for determining the representative prediction is compared to the commonly used MEAN and MEDIAN and evaluated on the real-life dataset Concrete.

\subsection{Dataset}
\label{section_datasets}

For comparison of the statistical analysis with real-life data, the regression data set \textbf{Concrete}~\cite{concrete} aquired from the data set platform Kaggle was used. The task is to estimate the concrete compressive strength as a highly nonlinear function of age and ingredients.

There are 1030 measurements provided and the given dataset is divided into $90\%$ Trainingdata and $10\%$ Testdata. A comparison regarding $R^2$, $RMSE$, $MAPE$ and $MAE$ to other approaches in the literature is made (detailed information in~\cite{concreteComp9}).

\subsection{Evaluation of the Comparative Analysis}
\label{subsection_evaluation_of_comarative_analysis}

The results of the comparison to MEAN and MEDIAN regarding $R^2$, $RMSE$, $MAPE$ and $MAE$ using the Concrete dataset are shown in Table~\ref{table_methodComparison}. The prediction errors with BS performing best with the highest $R^2$ of $0.934$ to $0.928$ with MEDIAN and $0.921$ with MEAN. Additionally all calculated error values are the lowest with BS at $RMSE=4.52$, $MAPE=10.8$ and $MAE=3.30$ and the highest with MEAN at $RMSE=5.64$, $MAPE=16.6$ and $MAE=4.49$. 
The evaluation shows that the results of MEDIAN at $RMSE=4.52$, $MAPE=12.6$ and $MAE=3.66$ are slightly worse, but comparable to the results of BS, which outperforms MEAN and MEDIAN in all categories.

\aptLtoX{\begin{table}
    \centering
\caption{Comparison to MEAN and MEDIAN. The best results are highlighted in bold.}
    \begin{tabular}{|c|c|c|c|c|}
    \hline
{\textbf{Method}} & {$R^2$} & {$RMSE$} & {$MAPE$} & {$MAE$}\\
    \hline
    $\Tilde{y}_{MEAN}$ & 0.921 & 5.64  & 16.6   & 4.49  \\
    $\Tilde{y}_{MEDIAN}$ & 0.928 & 4.89  & 12.6   & 3.66  \\
    $\Tilde{y}_{BS}$ & \textbf{0.934} & \textbf{4.52}  & \textbf{10.8}   & \textbf{3.30}  \\
    \hline
    \hline
     \end{tabular}
\label{table_methodComparison}
\end{table}}{\begin{table}[ht]
    \centering
\caption{Comparison to MEAN and MEDIAN. The best results are highlighted in bold.}
    \resizebox{0.96\columnwidth}{!}{%
    \begin{tabular}{|c|c|c|c|c|}
    \hline
\thead{\textbf{Method}} & \thead{\textbf{$R^2$}} & \thead{\textbf{$RMSE$}} & \thead{\textbf{$MAPE$}} & \thead{\textbf{$MAE$}}\\
    \hline
    $\Tilde{y}_{MEAN}$ & 0.921 & 5.64  & 16.6   & 4.49  \\
    $\Tilde{y}_{MEDIAN}$ & 0.928 & 4.89  & 12.6   & 3.66  \\
    $\Tilde{y}_{BS}$ & \textbf{0.934} & \textbf{4.52}  & \textbf{10.8}   & \textbf{3.30}  \\
    \hline
    \hline
     \end{tabular}
       }
\label{table_methodComparison}
\end{table}}

Table~\ref{table_concreteComparison} shows a comparison regarding $R^2$, $RMSE$, $MAPE$ and $MAE$ to other approaches in the literature acquired from~\cite{concreteComp9}. It shows that working with the combination of Alternating Transferfunctions~\cite{seitz2023alternating}, bagging and the new method of determining $\tilde{y}_{BS}$ leads to a top three ranking in each of the calculated error values of the compared literature without using any optimization or feature selection technique, except of $R^2$, which places the approach midfield.

\aptLtoX{\begin{table}[ht]
    \centering
\caption{Comparison to the literature with Concrete dataset. The best results are highlighted in bold.}
    \begin{tabular}{|c|c|p{90mm}|c|c|c|c|}
    \hline
{\textbf{Year}} & {\textbf{Ref.}} & {\textbf{Model}} &{$R^2$} & {$RMSE$} & {$MAPE$} & {$MAE$}\\
    \hline
1998  & \cite{concreteComp1} & ANN with manual optimization. & 0.91 & –     & –     & –     \\
2010  & \cite{concreteComp2} & Multiple additive regression tree with manual optimization. & 0.95 & 4.95  & 13.89 & –     \\
2013  & \cite{concreteComp3} & Wavelet gradient boosted ANN with manual optimization. & 0.95 & 5.75  & –     & 4.83  \\
2013  & \cite{concreteComp4} & ANN without optimization. & 0.93 & 6.33  & 15.3  & 4.41  \\
2014  & \cite{concreteComp5} & Ensemble stacking without tuning. & –    & 5.08  & 11.97 & 3.52  \\
2014  & \cite{concreteComp6} & Genetic weighted pyramid operation tree with manual optimization. & –    & 6.379 & 16.1  & 4.79  \\
2015  & \cite{concreteComp7} & Support vector regression with firefly optimization. & 0.87 & 4.86  & 9.81  & –     \\
2019  & \cite{concreteComp8} & Random forest without optimization. & 0.97 & 4.43  & 11.79 & 3.11  \\
2020  & \cite{concreteComp9} & XGBoost with automated feature selection and hyperparameter optimization. & \textbf{0.98} & \textbf{2.65}  & \textbf{7.4}   & \textbf{1.89}  \\
2025  & This & $\tilde{y}_{BS}$ of net ensemble predictions with Alternating Transferfunctions~\cite{seitz2023alternating}. & 0.93 & 4.52  & 10.8   & 3.30  \\
    \hline
    \hline
     \end{tabular}
\label{table_concreteComparison}
\end{table}
}{\begin{table}[ht]
    \centering
\caption{Comparison to the literature with Concrete dataset. The best results are highlighted in bold.}
    \resizebox{0.96\columnwidth}{!}{%
    \begin{tabular}{|c|c|p{90mm}|c|c|c|c|}
    \hline
\thead{\textbf{Year}} & \thead{\textbf{Ref.}} & \thead{\textbf{Model}} & \thead{\textbf{$R^2$}} & \thead{\textbf{$RMSE$}} & \thead{\textbf{$MAPE$}} & \thead{\textbf{$MAE$}}\\
    \hline
1998  & \cite{concreteComp1} & ANN with manual optimization. & 0.91 & –     & –     & –     \\
2010  & \cite{concreteComp2} & Multiple additive regression tree with manual optimization. & 0.95 & 4.95  & 13.89 & –     \\
2013  & \cite{concreteComp3} & Wavelet gradient boosted ANN with manual optimization. & 0.95 & 5.75  & –     & 4.83  \\
2013  & \cite{concreteComp4} & ANN without optimization. & 0.93 & 6.33  & 15.3  & 4.41  \\
2014  & \cite{concreteComp5} & Ensemble stacking without tuning. & –    & 5.08  & 11.97 & 3.52  \\
2014  & \cite{concreteComp6} & Genetic weighted pyramid operation tree with manual optimization. & –    & 6.379 & 16.1  & 4.79  \\
2015  & \cite{concreteComp7} & Support vector regression with firefly optimization. & 0.87 & 4.86  & 9.81  & –     \\
2019  & \cite{concreteComp8} & Random forest without optimization. & 0.97 & 4.43  & 11.79 & 3.11  \\
2020  & \cite{concreteComp9} & XGBoost with automated feature selection and hyperparameter optimization. & \textbf{0.98} & \textbf{2.65}  & \textbf{7.4}   & \textbf{1.89}  \\
2025  & This & $\tilde{y}_{BS}$ of net ensemble predictions with Alternating Transferfunctions~\cite{seitz2023alternating}. & 0.93 & 4.52  & 10.8   & 3.30  \\
    \hline
    \hline
     \end{tabular}
       }
\label{table_concreteComparison}
\end{table}}

\section{Conclusion and Further Research}
\label{section_conclusion_and_further_research}

With the presented method for the determination of the representative value $\Tilde{y}_{BS}$ using KDE, better results could be achieved in the analytical comparison than with conventional statistical methods, MEAN and MEDIAN.

Working with the combination of Alternating Transferfunctions~\cite{seitz2023alternating}, bagging and the new method provides comparable training results to known top approaches in the literature without having to use any complex optimization or feature selection techniques.

In terms of expected deviations,~\cite{friedman2007bagging} has shown several times that the variance in the data is not larger than the variance of the ensembles predictions. With a post training analysis the BS can be used to create a function for estimating the prediction error and determining expected deviation from the prediction. We already found out that high BSs typically result in low prediction errors. Further work on this might allow a significant reduction of estimated deviation compared to using the variance of the ensemble predictions.

Initial investigations under the availability of extremely sparse data have shown that BS also performs well for this. A more detailed look at this will certainly make another contribution to the literature. 
The question arises how useful the use of the BS is for tasks of classification.

While the Bagging Score is applied after training the NN ensemble, methods like cross-validation, to tune hyperparameters or to search for alternative ML models, or methods for analyzing the dataset like out-of-bag score~\cite{mohandoss2021outlier} might help to increase the prediction accuracy as well.

\begin{acks}

The authors gratefully acknowledge the \emph{Bayerisches Staatsministerium für Wirtschaft, Landesentwicklung und Energie} for funding for supporting the project \emph{LeMO2n - Lernende Multi-Skalen-Optimierung für SiO2-basierende Anodenmaterialien} under Grant no. 0703/68362/298/21/16/22/17/23/18/24; and Hightech Agenda Bayern.

\end{acks}

\section*{Disclosure Statement}
The authors report there are no competing interests to declare.

\section*{Data Availability Statement}
The data that support the findings of this study is openly available at https://kaggle.com (also see~\cite{concrete}). 


\begin{thebibliography}{4}

\bibitem{hunter2012selection}
Hunter, David, Hao Yu, Michael S Pukish III, Janusz Kolbusz, and Bogdan M Wilamowski. 2012. “Selection of proper neural network sizes and architectures—A comparative study.” IEEE Transactions on Industrial Informatics 8 (2): 228–240.

\bibitem{sykes1993introduction}
Sykes, Alan O. 1993. “An introduction to regression analysis.” .

\bibitem{breiman1996bagging}
Breiman, Leo. 1996. “Bagging predictors.” Machine learning 24: 123–140.

\bibitem{schapire2003boosting}
Schapire, Robert E. 2003. “The boosting approach to machine learning: An overview.” Nonlinear estimation and classification 149–171.

\bibitem{drucker1994boosting}
Drucker, Harris, Corinna Cortes, Lawrence D Jackel, Yann LeCun, and Vladimir Vapnik. 1994. “Boosting and other machine learning algorithms.” In Machine Learning Proceedings 1994, 53–61. Elsevier.

\bibitem{wang2010high}
Wang, Ying, Yong Fan, Priyanka Bhatt, and Christos Davatzikos. 2010. “High-dimensional pattern regression using machine learning: from medical images to continuous clinical variables.” Neuroimage 50 (4): 1519–1535.

\bibitem{bauer1999empirical}
Bauer, Eric, and Ron Kohavi. 1999. “An empirical comparison of voting classification algorithms: Bagging, boosting, and variants.” Machine learning 36: 105–139.

\bibitem{friedman2007bagging}
Friedman, Jerome H, and Peter Hall. 2007. “On bagging and nonlinear estimation.” Journal
of statistical planning and inference 137 (3): 669–683.

\bibitem{chen2009bagging}
Chen, Tao, and Jianghong Ren. 2009. “Bagging for Gaussian process regression.” Neurocomputing 72 (7-9): 1605–1610.

\bibitem{grandvalet2004bagging}
Grandvalet, Yves. 2004. “Bagging equalizes influence.” Machine Learning 55: 251–270.

\bibitem{guo2022predicting}
Guo, Hongwei, Xiaoying Zhuang, Jianfeng Chen, and Hehua Zhu. 2022. “Predicting earthquake-induced soil liquefaction based on machine learning classifiers: A comparative multi-dataset study.” International Journal of Computational Methods 19 (08): 2142004.

\bibitem{lin2024comprehensive}
Lin, Shan, Zenglong Liang, Shuaixing Zhao, Miao Dong, Hongwei Guo, and Hong Zheng. 2024. “A comprehensive evaluation of ensemble machine learning in geotechnical stability analysis
and explainability.” International Journal of Mechanics and Materials in Design 20 (2): 331–352.

\bibitem{parzen1962estimation}
Parzen, Emanuel. 1962. “On estimation of a probability density function and mode.” The annals of mathematical statistics 33 (3): 1065–1076.

\bibitem{wkeglarczyk2018kernel}
Weglarczyk, Stanislaw. 2018. “Kernel density estimation and its application.” In ITM web of conferences, Vol. 23, 00037. EDP Sciences.

\bibitem{seitz2023alternating}
Seitz, Philipp, and Jan Schmitt. 2023. “Alternating Transfer Functions to Prevent Overfitting in Non-Linear Regression with Neural Networks.” Journal of Experimental \& Theoretical Artificial Intelligence 1–22.

\bibitem{concrete}
Yeh, I-C. 1998. “Modeling of strength of high-performance concrete using artificial neural networks.” Accessed: 2023-09-13, https://www.kaggle.com/datasets/maajdl/yeh-concret-data.




\bibitem{chen2017tutorial}
Chen, Yen-Chi. 2017. “A tutorial on kernel density estimation and recent advances.” Biostatistics \& Epidemiology 1 (1): 161–187.

\bibitem{sheather2004density}
Sheather, Simon J. 2004. “Density estimation.” Statistical science 588–597.

\bibitem{mohandoss2021outlier}
Mohandoss, Divya Pramasani, Yong Shi, and Kun Suo. 2021. “Outlier prediction using random forest classifier.” In 2021 IEEE 11th Annual Computing and Communication Workshop and Conference (CCWC), 0027–0033. IEEE.

\bibitem{concreteComp1}
YEH, I.-C. Modeling of strength of high-performance concrete using artificial neural networks. Cement and Concrete research, 1998, 28. Jg., Nr. 12, S. 1797-1808.

\bibitem{concreteComp2}
CHOU, Jui-Sheng, et al. Optimizing the prediction accuracy of concrete compressive strength based on a comparison of data-mining techniques. Journal of Computing in Civil Engineering, 2011, 25. Jg., Nr. 3, S. 242-253.

\bibitem{concreteComp3}
ERDAL, Halil Ibrahim; KARAKURT, Onur; NAMLI, Ersin. High performance concrete compressive strength forecasting using ensemble models based on discrete wavelet transform. Engineering Applications of Artificial Intelligence, 2013, 26. Jg., Nr. 4, S. 1246-1254.

\bibitem{concreteComp4}
CHOU, Jui-Sheng; PHAM, Anh-Duc. Enhanced artificial intelligence for ensemble approach to predicting high performance concrete compressive strength. Construction and Building Materials, 2013, 49. Jg., S. 554-563.

\bibitem{concreteComp5}
CHOU, Jui-Sheng, et al. Machine learning in concrete strength simulations: Multi-nation data analytics. Construction and Building materials, 2014, 73. Jg., S. 771-780.

\bibitem{concreteComp6}
CHENG, Min-Yuan; FIRDAUSI, Pratama Mahardika; PRAYOGO, Doddy. High-performance concrete compressive strength prediction using Genetic Weighted Pyramid Operation Tree (GWPOT). Engineering Applications of Artificial Intelligence, 2014, 29. Jg., S. 104-113.

\bibitem{concreteComp7}
PHAM, Anh-Duc; HOANG, Nhat-Duc; NGUYEN, Quang-Trung. Predicting compressive strength of high-performance concrete using metaheuristic-optimized least squares support vector regression. Journal of Computing in Civil Engineering, 2016, 30. Jg., Nr. 3, S. 06015002.

\bibitem{concreteComp8}
HAN, Qinghua, et al. A generalized method to predict the compressive strength of high-performance concrete by improved random forest algorithm. Construction and Building Materials, 2019, 226. Jg., S. 734-742.

\bibitem{concreteComp9}
CHAKRABORTY, Debaditya; AWOLUSI, Ibukun; GUTIERREZ, Lilianna. An explainable machine learning model to predict and elucidate the compressive behavior of high-performance concrete. Results in Engineering, 2021, 11. Jg., S. 100245.





\end{thebibliography}
\end{document}